\pdfoutput=1

\documentclass[11pt]{article}
\usepackage{longtable}
\usepackage[margin=1in]{geometry} 
\usepackage{graphicx}
\usepackage[shortlabels]{enumitem}
\usepackage{tabularx}
\usepackage{array}
\usepackage{cuted}  
\usepackage{caption}    
\usepackage[preprint]{acl}

\usepackage{times}
\usepackage{latexsym}

\usepackage{amsmath}

\usepackage[T1]{fontenc}

\usepackage[utf8]{inputenc}

\usepackage{microtype}

\usepackage{inconsolata}

\usepackage{graphicx}

\usepackage{booktabs}       
\usepackage{enumitem}       
\usepackage{amssymb}        
\usepackage[hang,flushmargin]{footmisc}  

\usepackage{multirow}
\usepackage{tabularray}
\usepackage{lipsum}
\usepackage{stfloats}

\title{TopicProphet: Prophesies on Temporal Topic Trends and Stocks}

\author{Olivia Kim \\
  Emory CS Department / Atlanta GA \\
  \texttt{olivia.kim@emory.edu}}

\begin{document}
\maketitle

\begin{abstract}
Stocks can’t be predicted. Despite many hopes, this premise held itself true for many years due to the nature of quantitative stock data lacking causal logic along with rapid market changes hindering accumulation of significant data for training models. To undertake this matter, we propose a novel framework, TopicProphet, to analyze historical eras that share similar public sentiment trends and historical background. Our research deviates from previous studies that identified impacts of keywords and sentiments - we expand on that method by a sequence of topic modeling, temporal analysis, breakpoint detection and segment optimization to detect the optimal time period for training. This results in improving predictions by providing the model with nuanced patterns that occur from that era’s socioeconomic and political status while also resolving the shortage of pertinent stock data to train on. Through extensive analysis, we conclude that TopicProphet produces improved outcomes compared to the state-of-the-art methods in capturing the optimal training data for forecasting financial percentage changes.

\end{abstract}
\section{Introduction}
\label{sec:introduction}
The stock market is often characterized as the epitome of unpredictability due to its nature of complexity embedded within the trading of stocks, as the very action greatly depends on the economy that experiences frequent price fluctuations and volatility imposed on by external factors. Being formed amidst the poker game of individuals, industries, and now even computers \cite{PATEL20152162}, trading patterns are full of capriciousness and uncertainty, which is amplified even further with the interplay of global economy, political policies, investor sentiments, and more. Stock prices reflect all available information that has the slightest degree of pertinence \cite{ca63b0b1-cdd9-30df-82da-7f0a29b60d9c}.
\\
The research on utilizing Artificial Intelligence models to predict market movements has been the topic of interest for an extended period of time because of its unique multilayered problem composition. Having the trading pace more rapid than ever, the difficulty of an individual to make a correct prediction towards stock trading is rising. Many models have been implemented to serve this purpose, such as Backtrader\cite{9115114}, QuantConnect\cite{taye2021trading}, PyAlgoTrade\cite{taye2021trading}, and Zipline\cite{pik2021hands}.\\
Despite great advancements made towards this field, the problem still remains as a challenge due to the passive nature of quantitative stock data, greatly limiting the causal reasoning inferable from it - its behaviours almost seem irrational \cite{lo2004adaptive}. Various machine learning models have failed to analyze such patterns to detect patterns and accurately simulate the dynamic movement of the stock market, proving the ineffectiveness of quantitative price data in estimating market practices and resulting actions yet emphasizing the importance of comprehending the behaviour of the traders. As such activities are greatly influenced by their surrounding environment, it is crucial to study the correlation between an epoch's spatiotemporal situation formulated from key events such as enacted political policies, economic crises, or even innovations and its resulting stock outcomes.\\
Such conclusions led to an increased interest in experimenting with various data types in addition to stock price data, among which several prominent studies showed results that reveal promising relations between news and stocks \cite{li2020multimodal}. This has been further substantiated by recent discoveries. While the previous common idea was that stocks depend on relations between companies\cite{feng2019temporal}, Hao et al. overturned this through demonstrating the great influence public sentiment has on the market \cite{hao2021retail}.

However, these methods also have limitations that prevent further advancement towards solving the conundrum. The novelty of our study lies within our attempts to tackle this issue of training data through “period optimization”. While analyzing public sentiment through news articles and social media play a powerful role in capturing the reasoning, there are more nuanced reasoning and patterns that aren’t captured. \\
To incorporate the hidden patterns induced from the socioeconomic situation of that era within the model’s training data, we present a new framework TopicProphet, which selects the optimal time period to train the model based on topic trends within news articles. We start by clustering the keywords by semantic similarity, which is then followed by prompting LLMs to identify the topic each cluster represents. We then create a time series of a topic’s trend, enabling us to analyze the temporal behaviour of each topics’ keyword frequencies. Subsequently, our algorithm selects a topic to conduct breakpoint analysis through detecting change points within its temporal trend, and segments the initial training dataset based on those breakpoints. Lastly, we do automatic segment selection that selects the optimal training segment to utilize for training based on the trend similarities of topics. Experiment results indicate that TopicProphet produces superior predictive results and therefore validates the effectiveness of periodic data in capturing nuanced relationships.

\section{Related Work}
Using the proposed method, our goal is to effectively segment financial data using topic trends of political and economic U.S. news and choose the most relevant segment for model prediction. 
\subsection{Stock Price Prediction Models}
Numerous predictive models have been developed to forecast stock trends, however all models exhibit limitations as it is impossible to predict the future with 100\% certainty. Various predictive models have include factor investing models such as Fama French\cite{MOSOEU202255}, statistical methods such as the Facebook Prophet Model\cite{DASH202169}, deep learning models such as LSTM networks with transformer models like FinBERT\cite{hossain2024finbertbilstmdeeplearningmodel}, and various LLMs\cite{abe2024leveraginglargelanguagemodels} such as GPT-4\cite{openai2024gpt4technicalreport}.\\
To address the limitations of quantitative models, many have tried manipulating the training data given to predictive models to impact the accuracy of the results. Event driven insights from LLMs\cite{kurisinkel2024text2timeseriesenhancingfinancialforecasting}, news based day trading strategies employing reinforcement learning\cite{sawhney-etal-2021-quantitative}, and topological data analysis\cite{huang2024exploringapplicationstopologicaldata} have shown improvements in prediction accuracy. The time that the model is trained on, as well as the time period that the model is predicting has an impact on the performance metrics of the model\cite{fons2024evaluatinglargelanguagemodels}. 

\subsection{Topic Modeling}
To find patterns in news topics, topic modeling can be used to identify clusters of similar words in text. Latent Dirichlet Allocation (LDA) being one of the most common approaches\cite{jelodar2018latentdirichletallocationlda}. Using keywords that are predetermined by the author is beneficial to topic modeling as word-frequency is often not forward looking\cite{LU2021102594}. TopicGPT\cite{pham2024topicgptpromptbasedtopicmodeling} built upon the LDA framework and relies on prompting an LLM to generate topics. This method produces higher quality, more interpretable topics then prior clustering methods.\\
Topic modeling can also be used to uncover the evolution of topics in data\cite{VAHIDNIA2024123279}. Topic frequencies in news events hold patterns, which can be used to track certain trends\cite{hu2023learnpastevolvefuture}. Therefore topic frequencies can effectively identify periods of great volatility for a particular topic. By identifying these periods, we propose segmenting the training data for performance analysis on the prediction model.
\subsection{Breakpoint Detection}
To automate identifying critical periods, breakpoint detection can be used. Bayesian Online Changepoint Detection\cite{adams2007bayesianonlinechangepointdetection} has been used to predict significant market events using the Dow Jones Industrial Average.\\
The Pruned Extract Linear Time (PELT) method is optimal for longer data sets, where the length of the data set affects the number of breakpoints found. This approach is similar to the Optimal Partitioning approach\cite{Jackson_2005} but with a reduction in the computational cost that does not affect the quality of the result. Ruptures is a Python package for multiple change point detection, and contains various methods to do so\cite{truong2018ruptureschangepointdetection}. 
\subsection{Our Contribution}
Building upon TopicGPT’s methodology, we leverage generated topics using historical news data to increase the prediction accuracy in Facebook’s Prophet model. While news data has been used for model enhancement, specifically sentiment related prediction, we propose using generated news topic frequencies to further improve model accuracy by automating segmentation by performing breakpoint detection using the PELT model and selecting relevant training data.

\section{Approach}
\label{sec:approach}

Our framework to produce the optimal stock price percentage change consists of multiple stages: topic modeling, time period segmentation and selection, and lastly prediction.\\

\subsection{Topic Modeling}

\begin{figure}[h!]
    \centering
    \includegraphics[width=0.5\textwidth]{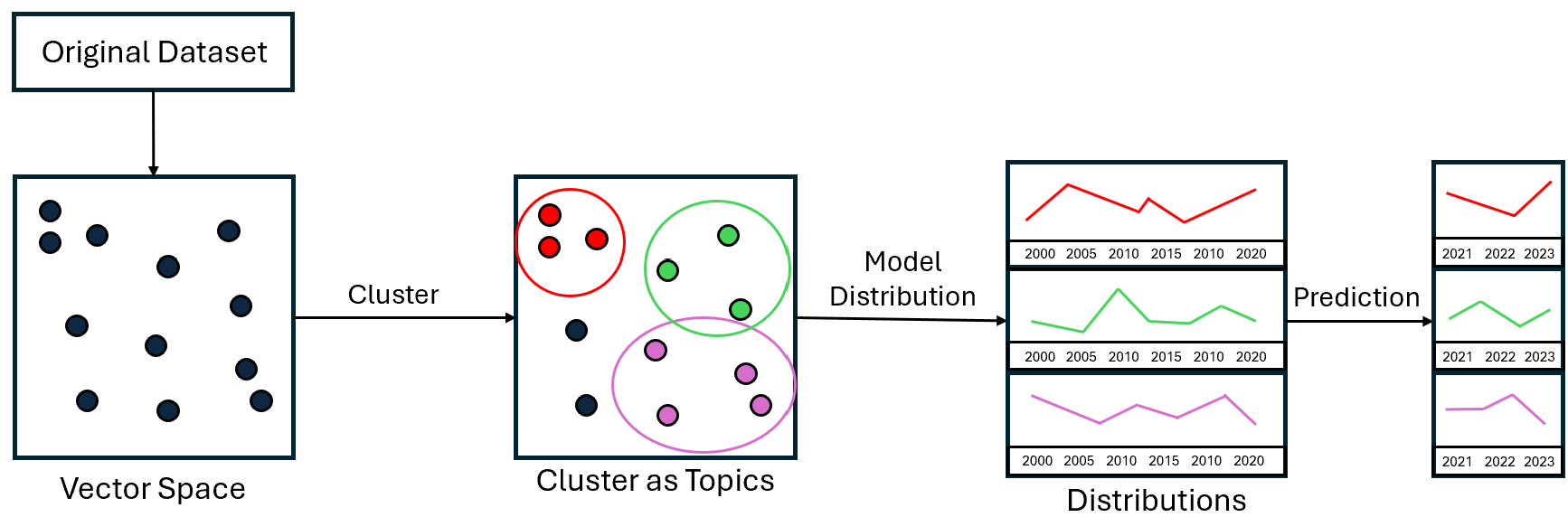} 
    \caption{Topic Modeling to Temporal Topic Trends}
    \label{fig:sample_image} 
\end{figure}
\subsubsection{Generate Keyword Embeddings}
Given the keywords extracted from each NYT article instance in the database $a_k$, we obtain the embedding vector $e_{a_k}$ generated using a sentence embedding model, where in our framework the BGE model\cite{bge_embedding} is selected. Contextual embeddings generated from such processes are stored to be later used in clustering.\\
\[e_{a_k} = SentenceEmbeddings(a_k)\]

\subsubsection{Dimensionality Reduction}
To amplify the effectiveness of clustering steps that are to be taken, dimensionality reduction using the UMAP (Uniform Manifold Approximation and Projection) technique is required\cite{mcinnes2020umapuniformmanifoldapproximation}. Research has proven UMAP’s efficacy in preserving the structural integrity of both the local and global data due to its scalability and non-linear dimensionality reduction processes, therefore demonstrating its suitability in this task compared to other methods like t-SNE or PCA\cite{MENG2024105887}. Given the embeddings of the keywords from each article, $e_{a_k}$, we are able to retrieve the dimensionality reduced embeddings, $r_{e_{ak}}$.
\[r_{e_{ak}} = UMAP(e_{a_k})\]

\subsubsection{Keyword Clustering}
The next step is done via the HDBSCAN (Hierarchical Density-Based Spatial Clustering of Applications with Noise) technique, which clusters the UMAP-reduced embeddings\cite{mcinnes2020umapuniformmanifoldapproximation}. Being able to capture complex data structures and non linear relationships due to its nature of being able to handle clusters of various shapes and densities, unlike traditional clustering algorithms like K-Means which assume all clusters are spherical and of the same size, HDBSCAN successfully facilitates the grouping of the data. Clustering based on density, the technique’s ability to automatically determine the number of clusters is crucial for this research. As a result, we are able to produce clusters, $c_i$, using the HDBSCAN upon the UMAP-reduced embeddings, $r_{e_{ak}}$.
\[c_i = HDBSCAN(r_{e_{ak}})\]

\subsubsection{Topic Generation}
The next step partakes in the recent development of LLM’s interpretation ability, proven by Pham et al.’s development of the TopicGPT framework. In order to fully automate the steps, the top 20 most frequent keywords per cluster generated from HDBSCAN were given to GPT-4, which was then prompted to generate a topic label that best represents the cluster. The generated topics and exemplary keywords that compose it can be seen in Appendix A.
\[t_i = ChatGPT(c_i)\]
As a result, we obtain topic label $t_i$ per cluster$c_i$.

\subsubsection{Topic Assignment}
Having generated topics per cluster, for further analysis to be done instances of the dataset must be structured by time and topic. In order to generate a time series depicting a topic’s trends in prominence, for each date we examine all keywords mentioned, then track which cluster of topics it belongs to. After labeling each article with its related topics, we demonstrate the topic’s prominence by frequency of that date. Each article consists of multiple articles, and each article may contain multiple topics. Based on the idea that the more articles about a certain topic is published that day the more relevant that topic is to the historical context of that day, a time series of each topic and their frequencies was constructed.

\subsection{Time Period Segmentation and Selection}
\subsubsection{Breakpoint Detection}
Changepoint / breakpoint detection in time series\cite{TRUONG2020107299} will be implemented in this paper to identify the “change points” in time when the trend of topics change significantly. While this can be conducted via the Prophet model, the “Ruptures”\cite{truong2018ruptureschangepointdetection} Library will be employed as it leverages a wide array of different breakpoint detection algorithms. The Pelt (Pruned Exact Linear Time) algorithm’s RBF (Radial Basis Function) model was chosen as it is able to minimize approximation errors while segmenting the time series. This is essential to our research as we desire to be able to detect breakpoints that not only were widely known of but also were hidden, unsupervised learning will be conducted without setting a definite number of breakpoints. While keeping the segmentation process automated, the minimum segment length will be predefined in order to avoid overfitting and over-segmentation.

\subsubsection{Segmentation}
Using the breakpoints retrieved, the training data will be segmented by every midpoint of breakpoints. As an example, if one breakpoint is 2001 and the other breakpoint is 2008, the training data will be segmented from 2000, the start of the data set, to 2005, the midpoint between the two break points, and the next segment will be from 2005 to the next midpoint. This method is employed to effectively capture the pre and post effects of the major event detected as a breakpoint.

\subsubsection{Segment Selection}
Based on our hypothesis that stock data itself lacks causal reasoning and that the reasoning behind prediction should be based on public sentiment and social context represented through topic trends, our true novelty of the study lies in this section of selecting the best training segment based on how closely the topic trends of the past resemble the predicted topic trends of the target era.
\begin{figure}
    \centering
    \includegraphics[width=\linewidth]{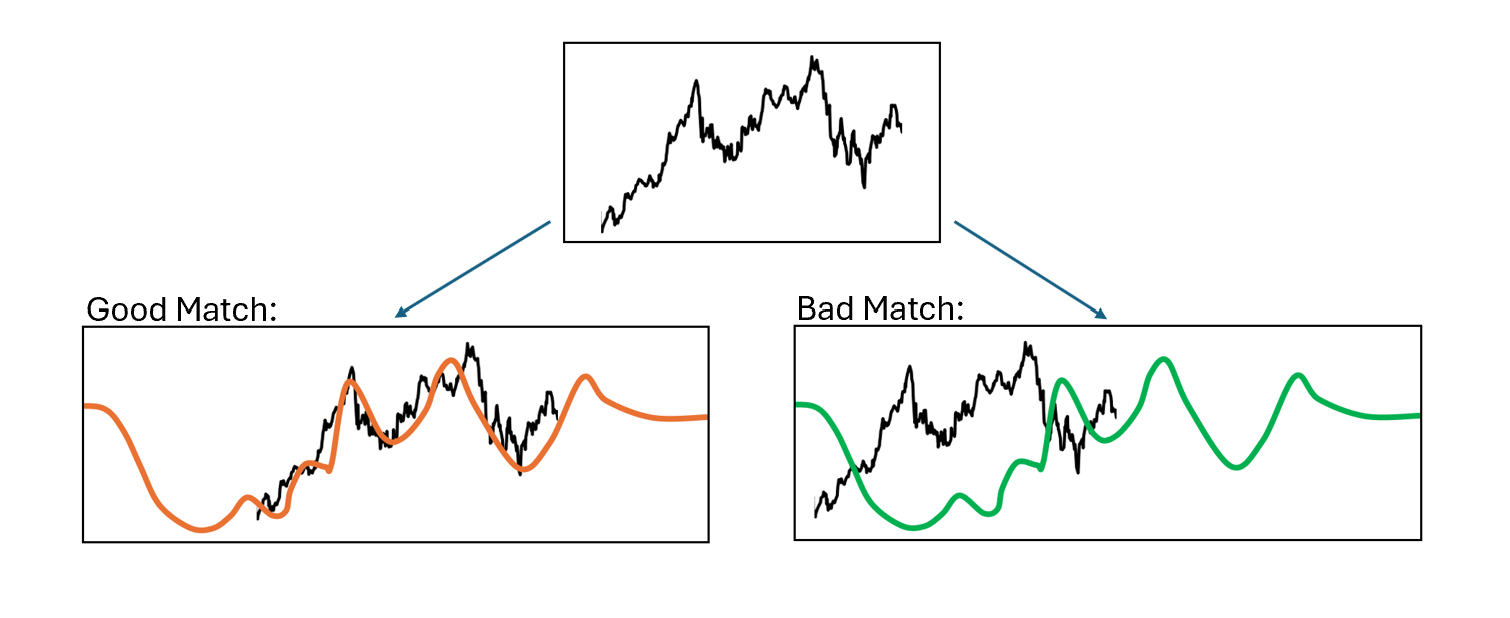}
    \caption{Exemplary Matching of Topic Trends}
    \label{fig:enter-label}
\end{figure}\\

\textbf{Predict Future Topic Trend} In order to do so, the Prophet model will be given all the training data and then used to forecast the future trends of topic frequencies for the target period, which is made possible because Prophet is an additive regressor model that has the capability to forecast most quantitative data.
\\
\textbf{Trend Similarity Metrics} To measure the similarity of a past time period with the future we want to predict, the segment's topic trend is compared with the predicted topic trend generated from the aforementioned method using the combination of three metrics: Pearson correlation, cosine similarity, and dynamic time warping.
\begin{itemize}
    \item Pearson correlation: Measures linear correlation between stock data and a singular topic trend data of the same time segment.
    \item Cosine similarity: Measures the similarity between two vectors using the cosine angle. Used to calculate significance of a topic.
    \item Dynamic Time Warping: compares two time series data sets, even if they're not aligned in time, by warping the time axis to find best trend alignment
\end{itemize}
Each of the metrics are found using equations: 
\[\mathcal{PW}_i = |Pearson(T_i,S)| \times (1-p)\]
\[\mathcal{CS}_i = 1 - \frac{T_i \cdot S}{||T_i|| \cdot||S||}\]
\[\mathcal{ND}_i = 1 - \frac{DTW(P_i, T_i)}{max(DTW(P_i,T_i),1)}\]
The two metrics Pearson Weight ($\mathcal{PW}$) and Cosine Similarity ($\mathcal{CS}$) are used to determine how significant the topic i ($T_i$) is by finding its correlation with the stock data of the same time period. The stocks with low correlation are dynamically mapped with a lower weight, as we want the significant topics to influence the results the most. These two metrics are then combined with the Normalized Dynamic Time Warping metric, which is the metric that actually conducts the matching between the past and the future topic trends. Each topic here is denoted with an index $i$ because this dynamic weighting is done by iterating through each topic and finding its effects individually.

\textbf{Combined Similarity Metric }
Finding the total similarity of the i-th topic trend of a certain time period is done by the equation below:
\[w_i = \alpha \cdot \mathcal{PW}_i + \beta \cdot \mathcal{ND}_i + \gamma \cdot \mathcal{CS}_i\]
where $w_i$ resembles the combined weight of each topic and coefficients $\alpha, \beta$, and $\gamma$ are calculated automatically based on the given dataset's characteristics. The three coefficients sum to a value of 1, weighting each metric differently: $\alpha$ scales differently per size of the data set, $\beta$ and $\gamma$ complementarily take up the remaining distribution depending on the ratio between the data set's trend strength and noise. All three coefficients are solved by fitting their independent variable into a logistic equation, which is used to scale the coefficients by data set. The calculated coefficients can be shown in appendix D.

\textbf{Finding Optimal Segment}
The final similarity score is calculated by summing the weights of all $N$ topics:
\[total = \sum_{i=1}^{N} w_i\]
Using this, the segments are ranked by the total score of each segment, making the segment with the lowest total score the most optimal segment to train the predictive model on.

\subsection{Stock Prediction}
Prediction will be done by Prophet, an additive regressor model specialized for time-series forecasting. Given training data, Prophet will return a continuous time series predicting the percentage change from the previous day. Again fortifying the notion that stock prediction's logic is to be based upon topic trends and not stock trends, time series of each topic's frequency trend will be added as a regressor.
\[\hat{y} = Prophet(S, T)\]
\begin{itemize}
    \item $\hat{y}$: stock percentage change from previous day
    \item $S$: stock time series of given time period
    \item $T$: $T = \{T_1, T_2,...,T_N\}$, set of all topic time series of given time period
\end{itemize}

\section{Experiments}
\label{sec:experiments}

\subsection{Data Preprocessing}

Two different datasets were used to each represent structured stock data and unstructured news article data. The data set of stock prices was accessed through Yahoo Finance via its api, and the data set of New York Times news articles was provided by the kaggle dataset “NYT Articles: 2.1M+ (2000-Present) Daily Updated”\cite{kaggle_dataset}. The dataset consists of over 2 million New York Times articles spanning from January 1st, 2000, to present and is updated daily. Each article is broken down to display these following key features:

\begin{itemize}
    \item Abstract: A summary of the article.
    \item Web url: A link to the article
    \item Headline: The title of the article.
    \item Keywords: Tags associated with the article.
\end{itemize}

The dataset was selected for use because of its large database of articles with the keywords extracted for each one. As our study focuses on trend analysis of topics using keywords of news articles, this data set was apt for the purpose of acquiring keywords.

Preprocessing was done in order to ensure quality and adaptability of the data. We removed instances that do not have a value in the keyword feature, which eliminated 22\% of the initial data set. The removal process was furthered as categories outside the scope of our research - our study focuses on political and economic events - were eliminated.
The last preprocessing step was to retrieve the keyword values only from the keyword feature, as it is formatted using a template of keyword types.

\begin{figure}[h!]
    \centering
    \includegraphics[width=0.5\textwidth]{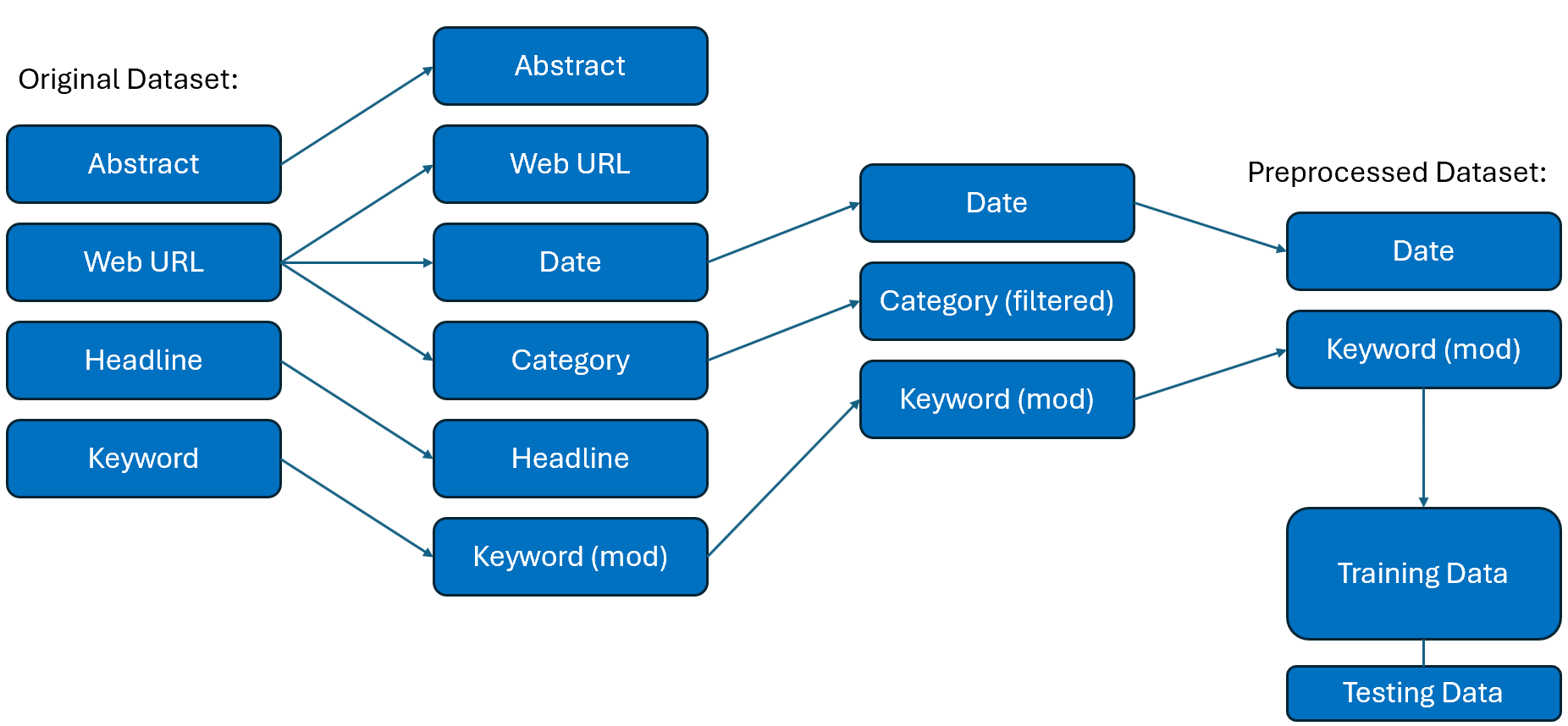} 
    \caption{Preprocessing Steps}
    \label{fig:sample_image} 
\end{figure}

The data was then separated into training/development data and testing data. To ensure that there was enough data to train the model for both segmented and unsegmented data, training data was set to include data from 2000-01-01 to 2018-05-10, validation 2018-05-10 to 2019-05-05, testing 2019-05-05 to 2024-05-01 and testing validation from 2024-05-01 to 2024-11-01. This split allocated 74\% of the data for training, 4\% for validation, 20\% for testing and 2\% for testing validation.
\begin{table}
\centering
\setlength{\arrayrulewidth}{0.5mm}
\setlength{\tabcolsep}{2pt}
\begin{tabular}{|l|l|l|l|l|}
\hline
 & Train & Train-Val&Testing &Test-Val\\
\hline
YFinance& 74\%& 4\%&20\%&2\%\\ \hline 
 NYT Articles & 74\%& 4\%& 20\%&2\%\\ \hline

\end{tabular}

\end{table}

\subsection{Baselines}
To make our experiment analyze the correlation between news article keywords’ topic trends with the US economy in a macro scale, the S\&P 500 index was used as the structured stock data to train and predict the model on. In order to evaluate the efficacy of utilizing topic trends in stock predictions, Facebook’s Prophet Model\cite{unknown} was the only model employed to not only minimize variance that is to be caused when having multiple models but also to fully embody the effects of having different types, lengths, and eras of data to train on.
\begin{itemize}
    \item \textbf{Baseline} given only quantitative stock data ranging across the entire period
    \item \textbf{Recent stock data} is given different lengths of stock data from most “recent” point of time
    \item \textbf{Topic trend data} is given all data on keyword frequencies to use as regressors
    \item \textbf{Segmented stock and topic data} is given the stock and keyword frequency data of a time period defined by breakpoint analysis of a trend.

\end{itemize}
All models utilize embeddings generated by the ‘bge-base-en-v1.5’ model, which were then dimensionally reduced using UMAP and then clustered using HDBSCAN, and finally assigned a topic using the ‘gpt-4-0613’ model. Our frameworks were implemented via PyTorch\cite{paszke2019pytorchimperativestylehighperformance}, Huggingface transformers\cite{wolf-etal-2020-transformers}, and scikit-learn\cite{pedregosa2018scikitlearnmachinelearningpython}.

\subsection{Evaluation Metrics}

As the framework of the study is majorly constructed by three operations - Topic Modeling, Breakpoint Analysis, Prediction - that each have different goals and purposes, each step requires its own set of evaluation metrics.

\subsubsection{Topic Modeling}

The same metrics utilized for the original TopicGPT\cite{pham2024topicgptpromptbasedtopicmodeling} were utilized to analyze how topically aligned by comparing the generated topic label a keyword with the ground truth, a method well verified through previous studies\cite{hoyle-etal-2022-neural}. Each news article instance was annotated with a most probable topic after considering its news headline title and keywords, which was set as ground truth. Using these annotations, the Purity and ARI of each topic cluster was measured.\\
\textbf{Purity}. The Harmonic mean of purity\cite{zhao2005criterion} and Inverse purity was utilized to assign each ground truth topic category to the cluster that has the highest F1 score\cite{article}. The score returns a rate between 0 to 1, where rates closer to 0 demonstrate randomness whereas rates closer 1 signify consistency.

\[
\text{Purity} = \frac{\sum_{ij} \binom{n_{ij}}{2} - \frac{\sum_i \binom{a_i}{2} \sum_j \binom{b_j}{2}}{\binom{n}{2}}}
{\frac{1}{2} \left[\sum_i \binom{a_i}{2} + \sum_j \binom{b_j}{2}\right] - \frac{\sum_i \binom{a_i}{2} \sum_j \binom{b_j}{2}}{\binom{n}{2}}}
\]
\begin{itemize}
    \item $\binom{x}{2}$: combination (choose 2 from x)
    \item $n_{ij}$: Elements in cluster $i$ of $C_1$ and cluster $j$ of $C_2$
    \item $a_i$: Row sum (total elements in cluster $i$ of $C_1$
    \item $b_j$: Column sum (total elements in cluster $j$ of $C_2$
    \item $n$: Total number of elements.
\end{itemize}

\textbf{Adjusted Rand Index}. The ARI utilizes Rand Index measurements’ ability to measure the similarity of pairs to describe the similarity of two clusters using probability. Rates again range from 0 to 1 where 0 shows randomness of assignment and 1 illustrates consistency\cite{vinh2009information}.
\[
\text{F-measure} = \frac{2 \cdot \text{Purity} \cdot \text{Inverse Purity}}{\text{Purity} + \text{Inverse Purity}}
\]
\[
\text{Purity} = \frac{1}{n} \sum_i \max_j n_{ij}
\]
\[
\text{Inverse Purity} = \frac{1}{n} \sum_j \max_i n_{ij}
\]
\begin{itemize}
    \item $C = \{C_1, C_2, ..., C_k\}$: Clusters
    \item $G = \{G_1, G_2, ..., G_m\}$: Ground-truth labels
    \item $n_{ij}$: Number of points in cluster $C_i$ that belong to ground-truth category $G_j$
\end{itemize}

\subsubsection{Change point detection}
The performances of change point detection were measured by comparing the detected change points with two sets of known reference events using precision, recall, and F-score. A detection is a true positive if it’s within the tolerance time window of the reference point, which are either from a set of US major event dates generated using ‘gpt-4-0613’ or are either from a set of breakpoints detected from the S\&P 500 stock data. The precision is computed by dividing true positives by all detections, while recall is computed by dividing true positives by total references, and the F-score is:
\[
F = \frac{2 \times Precision \times Recall}{Precision + Recall}
\]

\subsubsection{Stock Prediction}
Due to the characteristics of stock data, classification of stock trend predictions to be ‘up’(1), ‘stationary’(0), or ‘down’ (-1) led to the models producing misleading outcomes, as results varied greatly by what determines the trend to be stationary or not. To reduce this issues, models were to predict percentage changes from the previous day, and four metrics of Mean Absolute Error(MAE), Mean Squared Error (MSE), Root Mean Squared Error (RMSE), and R-Squared ($R^2$) were applied to evaluate each model’s performance.

\subsection{Experimental Settings}

As we only utilized the “topic assignment” technique of TopicGPT due to cost issues, we utilized the default UMAP algorithm\cite{mcinnes2020umapuniformmanifoldapproximation}. In order to control the number of topics generated so it’s not too redundant but still is specific enough, hyperparameter tuning led our study to settle with the parameters: \verb|n_components| = 5,   \verb|n_neighbors = 15|,  \verb|spread| = 1.5,  \verb|min_dist| = 0.01,  \verb|metric| = 'cosine'. By the final run, we were able to determine 114 clusters. Building upon these results, the clustering algorithm HDBSCAN\cite{10.1007/978-3-642-37456-2_14} was given hyperparameters of  \verb|min_cluster_size| = 200,  \verb|min_samples| = 18 to ensure each cluster isn’t too small.
Our experiments were conducted using the NVIDIA GeForce RTX 4050 GPU with a CPU of 2400MHz.

\subsection{Results}
Table 1 shows the results of the baseline compared to differently configured models. The approach of utilizing topic trend data instead of only stock data proves to be more effective: the baseline produced an MSE value of 5203.81 and the all topics model produced 4856.98. Utilizing breakpoint analysis on all topics for data segmentation showed improvements on certain time periods yet showed worse results on others, which is elaborated in Analysis.\\
From the results, we observe the following trends:
\begin{enumerate}
    \item Using topic trend data as regressors improves results from baseline. This result proves the effect of utilizing topic trends for stock prediction.
    \item Using segmented data of time periods determined from breakpoint analysis of certain topics’ trends show improved results for certain segments while showing comparably worse results for others. This result can be interpreted as evidence that predictions can be better made when utilizing data from time periods of similar socioeconomic circumstances.
    \item Using different topics to determine breakpoints and segmenting data based on such breakpoints yield different results, even for a similar time period of similar time span. This infers that each topic affects the prediction stock data differently.
\end{enumerate}

\begin{strip}
\renewcommand{\arraystretch}{1.8}
\captionof{table}{Comparison of exemplary configurations' performance in S\&P stock prediction}
\centering
\begin{tabular}{|l|l|l|l|l|l|} 
\hline
\multicolumn{2}{|c|}{Configuration} & MAE & MSE & RMSE & R2 \\ \hline
Stock & Full Span & 0.66 & 0.80 & 0.89 & -0.22 \\ \hline
Stock \& All Topics & 2020-11-10 to 2021-04-18 & 0.59 & 0.69 & 0.83 & -0.06 \\ \hline
Stock \& All Topics & 2021-04-19 to 2021-08-22 & 0.66 & 0.74 & 0.86 & -0.13 \\ \hline
Stock \& All Topics & 2021-08-23 to 2021-12-15 & 0.68 & 0.76 & 0.87 & -0.17 \\ \hline
Stock \& All Topics & 2021-12-16 to 2022-03-20 & 0.66 & 0.81 & 0.90 & -0.24 \\ \hline
Stock \& All Topics & 2022-03-21 to 2022-10-05 & 0.60 & 0.72 & 0.85 & -0.10 \\ \hline
Stock \& All Topics & 2022-10-06 to 2024-04-30 & 0.59 & 0.66 & 0.81 & -0.02 \\ \hline
\end{tabular}
\end{strip}

\section{Analysis}
\label{sec:analysis}

\subsection{Ablation Study}
We conducted an ablation study based on two control variables: training data time interval, and topic trends that were used as regressors. 

\subsubsection{Time Interval}
The influence of the training data’s time interval was tested by breaking down the problem further into two subproblems: how long is the data, and when is the data. The first subproblem was created due to the common conception that the only valid training data for stocks is the recent few months. In order to verify the role of recency in stock data, we compared giving different lengths of “recent” data and evaluated the outcomes. Not surprisingly, results were best when trained on a small amount of data (1-2 months) prior to the cut off date, yet this can not be trusted due to its high overfitting. Any duration beyond that develops insignificant results.\\
\begin{figure}
    \centering
    \includegraphics[width=1\linewidth]{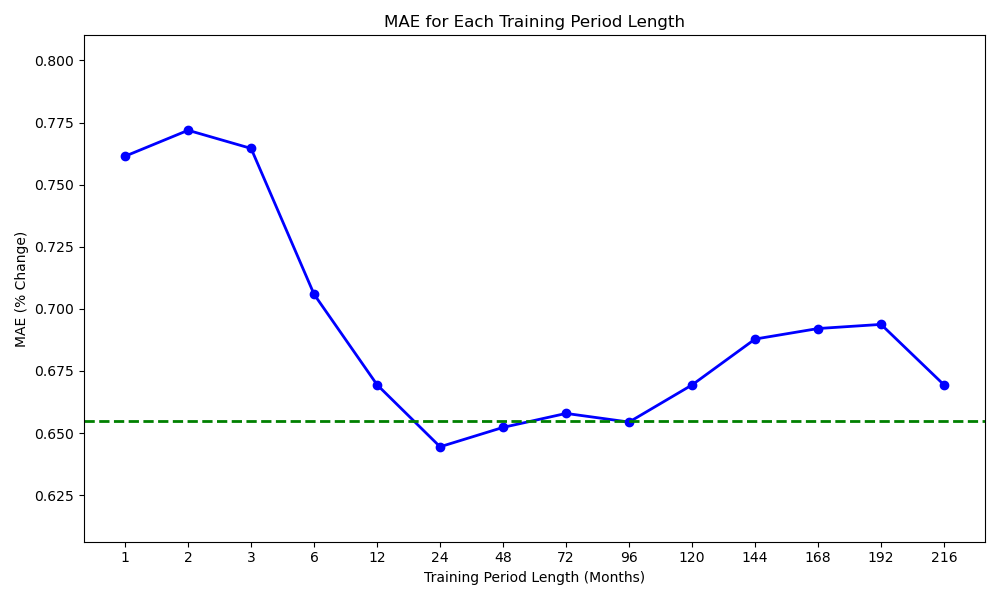}
    \caption{MAE of training period length of "Recent data"}
    \label{fig:enter-label}
\end{figure}
The second subproblem showcases the role of “when” the data is from. Stemming from our original hypothesis of the study, as stock data is influenced by many external factors, we believed that stock data of certain time periods will embody the socioeconomic and political influences within its data. In order to prove that hypothesis, we selected a few influential topics to perform breakpoint analyses and segment time periods based on the product. Evaluating all segments, certain topics performed considerably better in certain periods while also yielding worse results in other periods. The data in Table 1 was segmented using the topic “Business and Corporations”, retrieving best evaluations with data around 2008. This can be interpreted to the fact that both the training period of 2008, the Great Recession, and the testing period of late 2019 to late 2020, the Covid-19 Pandemic, were eras that experienced a substantial economic downfall. The stock data around 2008 well embodied that era’s socioeconomic instability and crises, therefore producing a good prediction for the also-unstable era of 2020.\\
The fact that the outcome using the optimized period (2008 in this case) returns less errors than only using a couple recent months of data (disregarding products of overfitting) allows us to see that while both data construct improved results due to its pertinence to the socioeconomic status quo, the segmented data embodies the external factors within its data more successfully.

\subsubsection{Topic Trends}
The contribution that each topic trend has in predicting stock data was measured minutely by conducting an ablation study of measuring the effect of withholding one topic at a time and seeing how the metrics alter when we predict stocks. While the subtraction of most topics produce a result similar to using all topics, certain topics lead to drastic differences, whether it's an improvement or not. \\
The topics that led to the model performing way worse were the topics gpt-4 labeled to be “miscellaneous”, covering multiple topics within itself. The removal of these miscellaneous dumps leading to such a notable downfall indicates that it works as the foundation for using topic trends as regressors, while also outlining a clear limitation of this study caused by our method of topic modeling.\\
On the contrary, there also were topics that improved results when the topic was removed from being a regressor. Interestingly enough, despite the media and entertainment industry being a part of the S\&P 500, most of the topics that improve performance when removed were related to media, television, and entertainment. Results per withheld topic is illustrated in figure 5.
\begin{figure*}
    \centering
    \includegraphics[width=\textwidth]{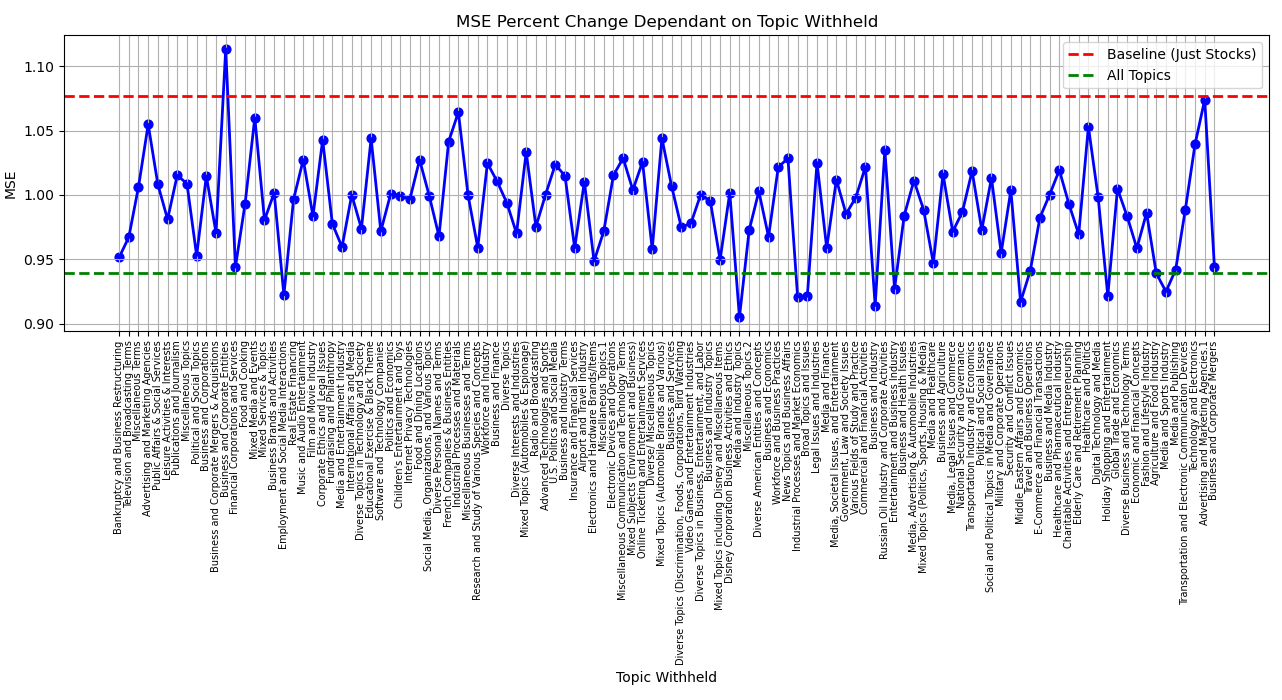}
    \caption{MSE percent change per withheld topic}
    \label{fig:enter-label}
\end{figure*}

\subsection{Error Analysis}
As error analysis, we examined the dates that returned an error, which was defined to be when the percent change difference (y-$\hat{y}$) of that date is greater than the median of the PCD, when using the baseline but then doesn’t output an error when using topic trend data as regressors. All cases that employ topic trend data are from the ablation study conducted above of observing outcomes of when each topic is withheld.\\
Whenever such an “error fix” is detected, we mark the date with the topic withheld. If a detected date is already marked, we increase the frequency each date was called.
After passing in the dates and the topics that affected such detection into GPT-4, we prompted it to return a significant world event that day and to choose a topic out of the given topics that event should be categorized to. The response returned that the dates of error fixes were most closely tied to topics “Business and Corporations”, “Software and Technology”, “U.S. Politics”, and more that were closely related to them.\\
These “error fixing” topics were then analyzed using change point detection to find its breakpoints and create time period segmentation from it,which were then used as individual training data. Results produced from such breakpoints demonstrated noteworthy improvement, which can be seen in the appendix.
\subsection{Discussions}
\subsubsection{Limitations}
We recognize four main limitations:
\begin{itemize}
    \item Although the prediction process heavily relies on the topic modeling done, our topic modeling was not optimal. There was a lot of overlap between clusters, and many clusters were just labeled as “mixed topics”, “miscellaneous”, “diverse topics” and therefore stopped further analyses and control methods using separations from being feasible.
    \item While we utilized the pre-extracted keywords given in the dataset due to resource limitations, many nuances each article had were lost in the process.
    \item Due to the topic assignment process being conducted by an LLM, the outcomes are variable and therefore are not able to be replicated precisely.
    \item There were no other models tested, so we cannot verify if the effectiveness of topic trends and using breakpoint analysis of “error fixing” topics to segment data would also apply to other models.
\end{itemize}

\subsubsection{Future Work}
Based on the limitations identified above, in the future, we plan to improve the experiment by being able to process the article directly for topic modeling using LLMs and automating the modeling and assignment process. Also, the research will also be conducted on different models to see if the effectiveness of trend analysis and using optimized segmented data persists across different circumstances. We will also attempt to conduct this process on LLMs as such models hold great potential in detecting nuanced and hidden patterns not easily detectable.
\section{Conclusion}
\label{sec:conclusion}
In this study, we analyze the outcomes of inputting different types of training data into the Facebook Prophet model in order to predict stock market trends with greater accuracy. The essence of this study lies in the attempt to optimize training data, which is to mitigate lack of causal relations within quantitative stock data by using topic trend data as regressors and also supplementing situational information through time period optimization via breakpoint analysis. Through rigorous testing and error analysis, TopicProphet proves its efficacy over the traditional methods of stock prediction, highlighting its novelty in providing missing nuances and therefore proving itself as a powerful tool for not only financial analysis and prediction, but also for socioeconomic trends. This research presents a method of systematic automation for prediction, once more highlighting the potency of combining LLMs with traditional models.

\cleardoublepage

\section{Appendix}
\label{sec:appendix}

\begin{strip}
 \renewcommand{\arraystretch}{1.8} 
    \centering
\textbf{Appendix A: Topics}\\
\captionof{table}{Example GPT Generated Cluster Topics and Representative Keywords}
    \begin{tabular}{|>{\raggedright\arraybackslash}p{5cm}|>{\raggedright\arraybackslash}p{10cm}|} 
    \hline 
             
Topics& Keywords\\ \hline 
             Bankruptcy and Business Restructuring& "['bankruptcies', 'exciteathome', 'amerco', 'panamsat', 'motient', 'nationsrent', 'mobilcom', 'threedo', 'ata', 'the', 'lyondellbasell', 'pathmark', 'amish', 'ampad', 'bahamas', 'series', 'gymboree']"\\ \hline 
             Television and Broadcasting Terms& "['television', 'itv', 'street', 'view', 'mole', 'intervideo', 'inhd', 'tlc', 'krups', 'kcet', 'simpsons', 'dottv', 'us', 'normal', 'itn', 'tashjian', 'joy', 'point', 'hotelevision', 'bienstock']"\\\hline
 Miscellaneous Terms& "['labor', 'dogs', 'unicor', 'imentor', 'ebeon', 'marchfirst', 'bax', 'freedrive', 'idealab', 'act', 'apw', 'hambrecht', 'equant', 'soccer', 'excelon', 'capitalia', 'leaves', 'convergys', 'guardent', 'aflac']"\\\hline
 Advertising and Marketing Agencies&"['advertising', 'interbrand', 'omnicom', 'foote', 'havas', 'frankel', 'howard', 'doremus', 'cmr', 'snuff', 'people', 'ecompanies', 'fast', 'esource', 'exercises', 'tlp', 'deepend', 'talk', 'bacara', 'amedia']"\\\hline
 Public Affairs \& Social Services& "['elections', 'handicapped', 'volunteers', 'publicaffairs', 'braskem', 'emotions', 'the', 'maps', 'peterson', 'poliomyelitis', 'photography']"\\\hline
 Leisure Activities \& Interests&"['retirement', 'television', 'xlibris', 'skiing', 'morgan', 'wines', 'pilots', 'coffee', 'series', 'twitter', 'art', 'basketball', 'whiskey']"\\\hline
 Publications and Journalism&"['newspapers', 'newsprint', 'redeye', 'translation', 'bugles', 'newspaperdirect', 'rumbo', 'blair', 'tageszeitung', 'men', 'color', 'reed', 'walter', 'rutgers', 'wikipedia', 'amish', 'berkeley', 'esquire', 'wpp', 'gristedes']"\\\hline
 Miscellaneous Topics&"['ethics', 'philosophy', 'informers', 'lying', 'soccer', 'watley', 'inkombank', 'tellium', 'pertamina', 'golf', 'lungs', 'tweedy', 'bonta', 'epicurum', 'football', 'kbr', 'armies', 'shaw', 'goldmarca', 'converium']" \\\hline
 Political and Social Topics&"['senate', 'kennedy', 'debating', 'flags', 'police', 'rusal', 'halloween', 'aged', 'morgan', 'amnesties', 'wolves', 'fences', 'assaults']"\\\hline
    \end{tabular}

\vspace{1cm}
    
    \textbf{Appendix B: Predictions based on different topics' breakpoints on S\&P 500}
  
    \captionof{table}{S\&P Baseline Results, All Topic Results, and Breakpoints Using All Topics}
    \begin{tabular}{|l|l|l|l|l|l|} 
    \hline 
             
 &Data Time Span& MAE
&MSE
&RMSE
&R2
 \\ \hline 
 Stock& 2000-01-01 2018-05-10
& 0.716424131& 1.076461677& 1.037526711&-0.212797653
\\ \hline 
 Stock \& All Topics
&  2000-01-01 2018-05-10& 0.654946527& 0.939530792& 0.969293966&-0.058524204\\\hline 
             Stock \& All Topics
&2000-01-01 2003-11-22
& 0.724832972&1.047634736&1.023540295&-0.180319724
\\ \hline 
             Stock \& All Topics&2003-11-23 2005-10-05
& 0.694644939&1.116516725&1.05665355&-0.257925751
\\\hline
  Stock \& All Topics
&2005-10-06 2006-11-20
& 0.73043101& 1.032281575& 1.016012586& -0.163022055
\\\hline
  Stock \& All Topics
&2006-11-21 2008-06-12
& 0.667182283& 1.031553622& 1.015654283& -0.162201905
\\\hline
  Stock \& All Topics
&2008-06-13 2010-01-30
& 0.815076384& 2.982161238& 1.726893523& -2.35985779
\\\hline
  Stock \& All Topics
&2010-01-31 2012-01-31
& 0.66390211& 0.942740476& 0.970948235& -0.0621404
\\ \hline 
  Stock \& All Topics
&2012-02-01 2015-05-08
& 0.662538961& 0.938377171& 0.9686987& -0.057224475
\\ \hline 
  Stock \& All Topics
&2015-05-09 2018-04-22
& 0.651496871& 0.932367746& 0.965591915& -0.050453944
\\ \hline 
 Stock \& All Topics& 2018-04-23 2018-05-09
& 0.670480402& 0.964148187& 0.981910478&-0.08625944
\\ \hline
    \end{tabular}
\captionof{table}{S\&P Breakpoints Using Business and Corporations Topic}
    \begin{tabular}{|l|l|l|l|l|l|} 
    \hline 
             
 &Data Time Span& MAE
&MSE
&RMSE
&R2
 \\ \hline 
 Stock& 2000-01-01 2018-05-10& 0.716424131& 1.076461677& 1.037526711&-0.212797653
\\ \hline 
 Stock \& All Topics& 2000-01-01 2018-05-10& 0.654946527& 0.939530792& 0.969293966&-0.058524204\\\hline \hline 
             Stock \& All Topics
&2000-01-01 2003-07-03
& 0.764218077&1.129528373&1.062792724&-0.272585351
\\ \hline 
             Stock \& All Topics
&
2003-07-04 2007-02-22
& 0.656753432&0.939277966&0.969163539&-0.058239357
\\\hline
  Stock \& All Topics
&2007-02-23 2010-06-29
& 0.724643864& 1.085475993& 1.041861792& -0.22295365
\\\hline
  Stock \& All Topics
&
2010-06-30 2011-07-17
& 0.668474968& 0.951636371& 0.975518514& -0.072162978
\\\hline
  Stock \& All Topics
&
2011-07-18 2012-08-20
& 0.687749538& 1.08362234& 1.040971825& -0.220865228
\\\hline
\textbf{  Stock \& All Topics}
&
\textbf{2012-08-21 2015-03-24}
&\textbf{ 0.653030032}& \textbf{0.930647964}&\textbf{ 0.964700971}& \textbf{-0.048516349}
\\ \hline 
  Stock \& All Topics
&
2015-03-25 2018-05-09
& 0.652942107& 0.937323337& 0.968154604& -0.05603717
\\\hline
    \end{tabular}

 \captionof{table}{S\&P Breakpoints Using Financial Corporations and Services Topic}
    \begin{tabular}{|l|l|l|l|l|l|} 
    \hline 
             
 &Data Time Span& MAE
&MSE
&RMSE
&R2
 \\ \hline 
 Stock& 2000-01-01 2018-05-10& 0.716424131& 1.076461677& 1.037526711&-0.212797653
\\ \hline 
 Stock \& All Topics& 2000-01-01 2018-05-10& 0.654946527& 0.939530792& 0.969293966&-0.058524204\\\hline \hline 
             Stock \& All Topics
&2000-01-01 2003-09-23
& 0.727180746&1.030490637&1.015130847&-0.161004291
\\ \hline 
             Stock \& All Topics
&
2003-09-24 2005-09-10
& 0.690445192&1.188251024&1.090069275&-0.33874534
\\\hline
  Stock \& All Topics
&2005-09-12 2007-01-01
& 0.72566944& 2.049891786& 1.431744316& -1.309514589
\\\hline
  Stock \& All Topics
&
2007-01-02 2008-03-31
& 0.71747926& 1.531633656& 1.237591878& -0.725618053
\\\hline
  Stock \& All Topics
&
2008-04-01 2018-05-09
& 0.664076436& 0.958063826& 0.978807349& -0.079404483
\\\hline
    \end{tabular}

\captionof{table}{S\&P Breakpoints Using Politics and Economics Topic}
    \begin{tabular}{|l|l|l|l|l|l|} 
    \hline 
             
 &Data Time Span& MAE
&MSE
&RMSE
&R2
 \\ \hline 
 Stock& 2000-01-01 2018-05-10& 0.716424131& 1.076461677& 1.037526711&-0.212797653
\\ \hline 
 Stock \& All Topics& 2000-01-01 2018-05-10& 0.654946527& 0.939530792& 0.969293966&-0.058524204\\\hline \hline 
             Stock \& All Topics
&2000-01-01 2003-06-05
& 0.737854525&1.058333892&1.028753562&-0.192373949
\\ \hline 
             Stock \& All Topics
&
2003-06-06 2008-11-16
& 0.683365432&0.992437225&0.996211436&-0.11813134
\\\hline
  Stock \& All Topics
&2008-11-17 2013-03-03
& 0.675696763& 0.972365982& 0.986086194& -0.095518035
\\\hline
  Stock \& All Topics
&
2013-03-04 2014-05-13
& 0.643933405& 0.919333884& 0.958819005& -0.035769318
\\\hline
  Stock \& All Topics
&
2014-05-14 2015-06-12
& 0.675975737& 0.944901752& 0.97206057& -0.064575407
\\\hline
  Stock \& All Topics
&
2015-06-13 2016-09-09
& 0.636648086& 0.905483978& 0.951569218& -0.020165294
\\ \hline 
  Stock \& All Topics
&
2016-09-10 2018-05-09
& 0.641185557& 0.908670525& 0.953242112& -0.023755424
\\\hline
    \end{tabular}

 \captionof{table}{S\&P Breakpoints Using Software and Technology Companies Topic}
    \begin{tabular}{|l|l|l|l|l|l|} 
    \hline 
             
 &Data Time Span& MAE
&MSE
&RMSE
&R2
 \\ \hline 
 Stock& 2000-01-01 2018-05-10& 0.716424131& 1.076461677& 1.037526711&-0.212797653
\\ \hline 
 Stock \& All Topics& 2000-01-01 2018-05-10& 0.654946527& 0.939530792& 0.969293966&-0.058524204\\\hline \hline 
             Stock \& All Topics
&2000-01-01 2008-06-19
& 0.756086341&1.545988395&1.243377817&-0.741790848
\\ \hline 
             Stock \& All Topics
&
2008-06-20 2010-03-24
& 0.721370204&1.142949945&1.069088371&-0.287706791
\\\hline
  Stock \& All Topics
&2010-03-25 2013-06-04
& 0.649179263& 0.916255627& 0.957212425& -0.032301194
\\\hline
  Stock \& All Topics
&
2013-06-05 2018-05-09
& 0.652980909& 0.931587146& 0.965187622& -0.04957448
\\\hline
    \end{tabular}

\captionof{table}{S\&P Final Results}
    \begin{tabular}{|l|l|l|l|l|l|} \hline

&Data Time Span
& MAE
&MSE
&RMSE
&R2
 \\ \hline  
             Stock
&Full Span
& 0.66
&0.80
&0.89&-0.22\\ \hline  
             Stock \& All Topics
&
2020-11-10 2021-04-18
& 0.59
&0.69
&0.83
&-0.06
\\ \hline 
  Stock \& All Topics
&2021-04-19 2021-08-22
& 0.66
& 
0.74
& 0.86
& -0.13
\\ \hline 
  Stock \& All Topics
&
2021-08-23 2021-12-15
& 
0.68
& 0.76
& 0.87
& -0.17
\\ \hline  
 Stock \& All Topics
& 2021-12-16 2022-03-20
& 0.66
& 
0.81
& 0.90
&-0.24\\ \hline  
 Stock \& All Topics
& 2022-03-21 2022-10-05
& 
0.60
& 0.72& 0.85&-0.10\\ \hline 
 Stock \& All Topics& 2022-10-06 2024-04-30& 0.59& 0.66& 0.81&-0.02\\ \hline
\end{tabular}

\vspace{1cm}

    \textbf{Appendix C: NASDAQ Prediction Results by Segment}
    \renewcommand{\arraystretch}{1.8} 
 \captionof{table}{Nasdaq  Baseline Results, All Topic Results, and Segmented Training Results}
    \begin{tabular}{|l|l|l|l|l|l|} 
    \hline 
             
 &Data Time Span& MAE
&MSE
&RMSE
&R2
 \\ \hline 
 Stock& Baseline& 0.890646372& 1.655228730& 1.286556928&-0.116176198\\ \hline 
 Stock \& All Topics& All Topics& 0.877676222& 1.618607745& 1.272245159&-0.091481441\\\hline \hline 
             Stock \& All Topics
&2000-01-01 2003-07-03
& 1.052145299
&2.019118117
&1.420956761
&-0.361559006
\\ \hline 
             Stock \& All Topics
&
2003-07-04 2007-02-22
& 0.86667809
&1.589597517
&1.260792416
&-0.071918872
\\\hline
  Stock \& All Topics
&2007-02-23 2010-06-29
& 0.934765554
& 1.759561483
& 1.326484633
& -0.18653127
\\\hline
  Stock \& All Topics
&
2010-06-30 2011-07-17
& 0.88097897
& 1.585469659
& 1.259154343
& -0.06913532
\\\hline
  Stock \& All Topics
&
2011-07-18 2012-08-20
& 0.862609148
& 1.564340144
& 1.250735841
& -0.054886979
\\\hline
\textbf{  Stock \& All Topics}
&
\textbf{2012-08-21 2015-03-24}
& \textbf{0.843497562}
& \textbf{1.511085023}
& \textbf{1.229261983}
& \textbf{-0.018975266}
\\ \hline 
  Stock \& All Topics
&
2015-03-25 2018-05-09
& 0.909252396
& 1.676003783
& 1.294605648
& -0.130185513
\\\hline
    \end{tabular}

 \captionof{table}{Nasdaq Baseline Results, All Topic Results, and Segmented Testing Results}
    \begin{tabular}{|l|l|l|l|l|l|} 
    \hline 
             
 &Data Time Span& MAE
&MSE
&RMSE
&R2
 \\ \hline 
 Stock& Baseline
& 0.422681047& 0.711225167& 0.843341667&-0.031353426\\ \hline 
 Stock \& All Topics& All Topics& 0.427739875& 0.708116730& 0.841496720&-0.026845856\\\hline \hline 
             Stock \& All Topics
&2019-11-01 2020-05-28& 0.420107983
&0.695683661
&0.834076532
&-0.008816561
\\ \hline 
             Stock \& All Topics
&
2020-05-29 2020-12-09& 0.430278891
&0.716140565
&0.846250888
&-0.038481284
\\\hline
  Stock \& All Topics
&2020-12-10 2021-08-02& 0.418118367
& 0.696080198
& 0.834314208
& -0.009391582
\\\hline
  Stock \& All Topics
&
2021-08-03 2021-12-15& 4.160766085
& 500.2707118
& 22.36673226
& -724.4466463
\\\hline
  Stock \& All Topics
&
2021-12-16 2022-03-18& 0.56736593
& 1.012760466
& 1.006360008
& -0.468612226
\\\hline
  Stock \& All Topics
&
2022-03-19 2022-10-05& 0.424078795
& 0.710536566
& 0.842933311
& -0.030354879
\\ \hline 
\textbf{  Stock \& All Topics}
&
\textbf{2022-10-06 2024-05-09}&\textbf{ 0.418445349}
&\textbf{ 0.693307954}
& \textbf{0.83265116}
& \textbf{-0.005371529}
\\\hline
    \end{tabular}

\vspace{1cm}

    \textbf{Appendix D: Russell 1000 Prediction Results by Segment}
    \renewcommand{\arraystretch}{1.8}
 \captionof{table}{Russell 1000 Baseline Results, All Topic Results, and Training Results}
    \begin{tabular}{|l|l|l|l|l|l|} 
    \hline 
             
 &Data Time Span& MAE
&MSE
&RMSE
&R2
 \\ \hline 
 Stock& Baseline
& 0.764263573& 1.266485894& 1.125382554&-0.079524666\\ \hline 
 Stock \& All Topics& All Topics& 0.754539663& 1.233275169& 1.110529229&-0.051216576\\\hline \hline 
             Stock \& All Topics
&2000-01-01 2003-07-03
& 0.817618902
&1.218374906
&1.103800211
&-0.244202267
\\ \hline 
             Stock \& All Topics
&
2003-07-04 2007-02-22
& 0.715284678
&1.017968227
&1.008944115
&-0.039547326
\\\hline
  Stock \& All Topics
&2007-02-23 2010-06-29
& 0.800138872
& 1.24190678
& 1.114408713
& -0.268232975
\\\hline
  Stock \& All Topics
&
2010-06-30 2011-07-17
& 0.7529549
& 1.106132062
& 1.051728131
& -0.129580076
\\\hline
  Stock \& All Topics
&
2011-07-18 2012-08-20
& 0.770843419
& 1.136892354
& 1.066251543
& -0.160992431
\\\hline
\textbf{  Stock \& All Topics}
&
\textbf{2012-08-21 2015-03-24}
& \textbf{0.703426124}
& \textbf{0.991993542}
&\textbf{ 0.995988726}
& \textbf{-0.013022024}
\\ \hline 
  Stock \& All Topics
&
2015-03-25 2018-05-09
& 0.742425813
& 1.063245749
& 1.031138084
& -0.085784651
\\\hline
    \end{tabular}

 \captionof{table}{Russell 1000 Baseline Results, All Topic Results, and Testing Results}
    \begin{tabular}{|l|l|l|l|l|l|} 
    \hline 
             
 &Data Time Span& MAE
&MSE
&RMSE
&R2
 \\ \hline 
 Stock& Baseline& 0.302376945& 0.348405303& 0.590258675&-0.074202268\\ \hline 
 Stock \& All Topics& All Topics& 0.307315096& 0.355773821& 0.596467787&-0.096920863\\\hline \hline 
             Stock \& All Topics
&2019-11-01 2020-05-28& 0.298755821
&0.328074651
&0.572778012
&-0.011518858
\\ \hline 
             Stock \& All Topics
&
2020-05-29 2020-12-09& 0.29072337
&0.338117379
&0.581478614
&-0.042482567
\\\hline
  Stock \& All Topics
&2020-12-10 2021-08-02& 0.29270355
& 0.324537148
& 0.56968162
& -0.000612037
\\\hline
  Stock \& All Topics
&
2021-08-03 2021-12-15& 0.365857968
& 0.419154701
& 0.647421579
& -0.292336614
\\\hline
  Stock \& All Topics
&
2021-12-16 2022-03-18& 0.329229298
& 0.403300569
& 0.635059501
& -0.243455199
\\\hline
  Stock \& All Topics
&
2022-03-19 2022-10-05& 0.301467811
& 0.356293133
& 0.596902951
& -0.098522003
\\ \hline 
  Stock \& All Topics
&
\textbf{2022-10-06 2024-05-09}& \textbf{0.28805881}
& \textbf{0.325801925}
& \textbf{0.570790614}
& \textbf{-0.004511595}
\\\hline
    \end{tabular}

\vspace{1cm}

    \textbf{Appendix E}
    \renewcommand{\arraystretch}{1.8} 
    \captionof{table}{Coefficients Per Data Set}
    \begin{tabular}{|l|l|l|l|l|l|l|l|l|} 
    \hline 
             
Dataset& Size&Volatility&Noise&Trend Strength & Periodicity& Pearson& DTW&Cosine\\ \hline 
             S\&P Validation&  4996
&1.0000002
&0.58
&0.66
& 0.06
& 0.60
& 0.19
&0.21
\\ \hline 
             S\&P Test&   1252
&1
&0.54
&0.70
& 0.06
& 0.20
& 0.22
&0.58
\\\hline
 Nasdaq Validation& 4617
& 1
& 0.67
& 0.55
& 0.07
& 0.60
& 0.29
&0.11
\\\hline
 Nasdaq Test& 1256
& 1
& 0.71
& 0.49
& 0.06
& 0.20
& 0.60
&0.20
\\\hline
 Russell 1000 Validation& 4643
& 1
& 0.50
& 0.75
& 0.06
& 0.60
& 0.04
&0.36
\\\hline
 Russell 1000 Test& 1130
& 0.9999999
& 0.62
& 0.62
& 0.06
& 0.17
& 0.52
&0.31
\\\hline
\end{tabular}

\vspace{1cm}
\textbf{Appendix F: Total Similarity Scores Per Segment}
    \renewcommand{\arraystretch}{1.8} 
    \captionof{table}{Ranking Per Training Segment for S\&P 500}
    \begin{tabular}{|l|l|l|l|} 
    \hline 
             
Segment& Start Date&End Date&Similarity Score\\ \hline 
             1&  2000-01-11&2003-07-03&19.230\\ \hline 
             2&   2003-07-07&2007-02-22&17.022\\\hline
 3& 2007-02-23& 2010-06-29& 9.704\\\hline
 4& 2010-06-30& 2011-07-15& 6.126\\\hline
 5& 2011-07-18& 2012--08-20& 6.087\\ \hline 
 \textbf{6}& \textbf{2012-08-21}& \textbf{2015-03-24}&\textbf{5.963}\\\hline
 7& 2015-03-25& 2018-05-09&7.302\\\hline
    \end{tabular}

    \captionof{table}{Ranking Per Testing Segment for S\&P 500}
    \begin{tabular}{|l|l|l|l|} 
    \hline 
             
Segment& Start Date&End Date&Similarity Score\\ \hline 
1&  2020-11-10&2021-04-16&3.519\\ \hline 
\textbf{2}&   \textbf{2021-04-19}&\textbf{2021-08-20}&\textbf{2.867}\\\hline
 3& 2021-08-23& 2021-12-15& 3.007\\\hline
 4& 2021-12-16& 2022-03-18& 3.355\\\hline
 5& 2022-03-21& 2022-10-05& 3.161\\ \hline 
 6& 2022-10-06& 2024-04-30&3.204\\\hline
    \end{tabular}

\captionof{table}{Ranking Per Training Segment for Nasdaq}
    \begin{tabular}{|l|l|l|l|} 
    \hline 
             
Segment& Start Date&End Date&Similarity Score\\ \hline 
             1&  2000-01-11&2003-07-03&6.596\\ \hline 
             2&   2003-07-07&2007-02-22&6.604\\\hline
 3& 2007-02-23& 2010-06-29& 3.286\\\hline
 4& 2010-06-30& 2011-07-15& 2.244\\\hline
 5& 2011-07-18& 2012--08-20& 1.880\\ \hline 
 \textbf{6}& \textbf{2012-08-21}& \textbf{2015-03-24}&1.739\\\hline
 7& 2015-03-25& 2018-05-09&2.502\\\hline
    \end{tabular}

    \captionof{table}{Ranking Per Testing Segment for Nasdaq}
    \begin{tabular}{|l|l|l|l|} 
    \hline 
             
Segment& Start Date&End Date&Similarity Score\\ \hline 
 1& 2019-11-01 & 2020-05-28& 2.561\\ \hline 
 2& 2020-11-10&2021-04-16&2.801\\ \hline 
 3& 2021-04-19& 2021-08-20& 2.470\\\hline
 4& 2021-08-23& 2021-12-15& 2.815\\\hline
 5& 2021-12-16& 2022-03-18& 2.422\\\hline
 6& 2022-03-21& 2022-10-05& 2.523\\ \hline 
 \textbf{7}& \textbf{2022-10-06}& \textbf{2024-04-30}&\textbf{2.417}\\\hline
    \end{tabular}

\captionof{table}{Ranking Per Training Segment for Russell 1000}
    \begin{tabular}{|l|l|l|l|} 
    \hline 
             
Segment& Start Date&End Date&Similarity Score\\ \hline 
             1&  2000-01-11&2003-07-03&14.524\\ \hline 
             2&   2003-07-07&2007-02-22&13.817\\\hline
 3& 2007-02-23& 2010-06-29& 7.449\\\hline
 4& 2010-06-30& 2011-07-15& 4.725\\\hline
 5& 2011-07-18& 2012--08-20& 4.379\\ \hline 
 \textbf{6}& \textbf{2012-08-21}& \textbf{2015-03-24}&4.210\\\hline
 7& 2015-03-25& 2018-05-09&5.663\\\hline
    \end{tabular}

    \captionof{table}{Ranking Per Testing Segment for Russell 1000}
    \begin{tabular}{|l|l|l|l|} 
    \hline 
             
Segment& Start Date&End Date&Similarity Score\\ \hline 
1&  2019-11-01&2020-05-28&1.000\\ \hline 
\textbf{2}&   2020-05-29&2020-12-09&0.999\\\hline
 3& 2020-12-10& 2021-08-02& 1.000\\\hline
 4& 2021-08-03& 2021-12-15& 0.999\\\hline
 5& 2021-12-16& 2022-03-18& 1.000\\ \hline 
 6& 2022-03-21& 2022-10-05&0.999\\\hline
 7& 2022-10-06& 2024-04-30&1.000\\\hline
    \end{tabular}

\vspace{1cm}
\textbf{Appendix G}
    \renewcommand{\arraystretch}{1.8} 
    \captionof{table}{Topic Modeling Metrics}
    \begin{tabular}{|l|l|} 
    \hline 
             
Purity& 0.500
\\ \hline 
             ARI&  0.042
\\ \hline 
             Precision&   0.203
\\\hline
 Recall& 0.130
\\\hline
 F1& 0.151\\\hline
    \end{tabular}

\end{strip}

\end{document}